\documentclass[letterpaper]{article} 
\usepackage{aaai2027}  
\nocopyright
\usepackage[hyphens]{url}  
\usepackage{graphicx} 
\usepackage{natbib}  
\usepackage{caption} 
\usepackage{algorithm}
\usepackage{algorithmic}
\usepackage{amsmath,amssymb,amsfonts}

\usepackage{newfloat}
\usepackage{listings}
\DeclareCaptionStyle{ruled}{labelfont=normalfont,labelsep=colon,strut=off} 
\floatstyle{ruled}
\newfloat{listing}{tb}{lst}{}
\floatname{listing}{Listing}

\usepackage{booktabs}
\usepackage{multirow}
\usepackage{makecell}
\usepackage{wasysym}
\usepackage{amssymb} 

\makeatletter
\def\@maketitle{%
  \def\aaai@theauthors{\if T\showauthors@on\@author\else Anonymous submission\fi}
  \newcounter{aaai@eqfn}\setcounter{aaai@eqfn}{0}%
  \newcounter{aaai@corrfn}\setcounter{aaai@corrfn}{0}%
  \newif\ifaaai@corrmulti\aaai@corrmultifalse
  \newsavebox{\titlearea}
  \sbox{\titlearea}{
    \let\footnote\relax\let\thanks\relax%
    \setcounter{footnote}{0}%
    \def\equalcontrib{%
      \ifnum\value{aaai@eqfn}=0%
        \footnote{These authors contributed equally.}%
        \setcounter{aaai@eqfn}{\value{footnote}}%
      \else%
        \footnotemark[\value{aaai@eqfn}]%
      \fi%
    }%
    \def\corresponding{%
      \ifnum\value{aaai@corrfn}=0%
        \footnote{Corresponding author.}%
        \setcounter{aaai@corrfn}{\value{footnote}}%
      \else%
        \footnotemark[\value{aaai@corrfn}]%
      \fi%
    }%
    \vbox{%
      \hsize\textwidth%
      \linewidth\hsize%
      \vskip 0.625in minus 0.125in%
      \centering%
      {\LARGE\bf \@title \par}%
      \vskip 0.1in plus 0.5fil minus 0.05in%
      {\Large{\textbf{\aaai@theauthors\ifhmode\\\fi}}}%
      \vskip .2em plus 0.25fil%
      {\normalsize \aaai@affiliations\ifhmode\\\fi}%
      \vskip 0.5em%
      \includegraphics[width=\textwidth,height=0.28\textheight,keepaspectratio]{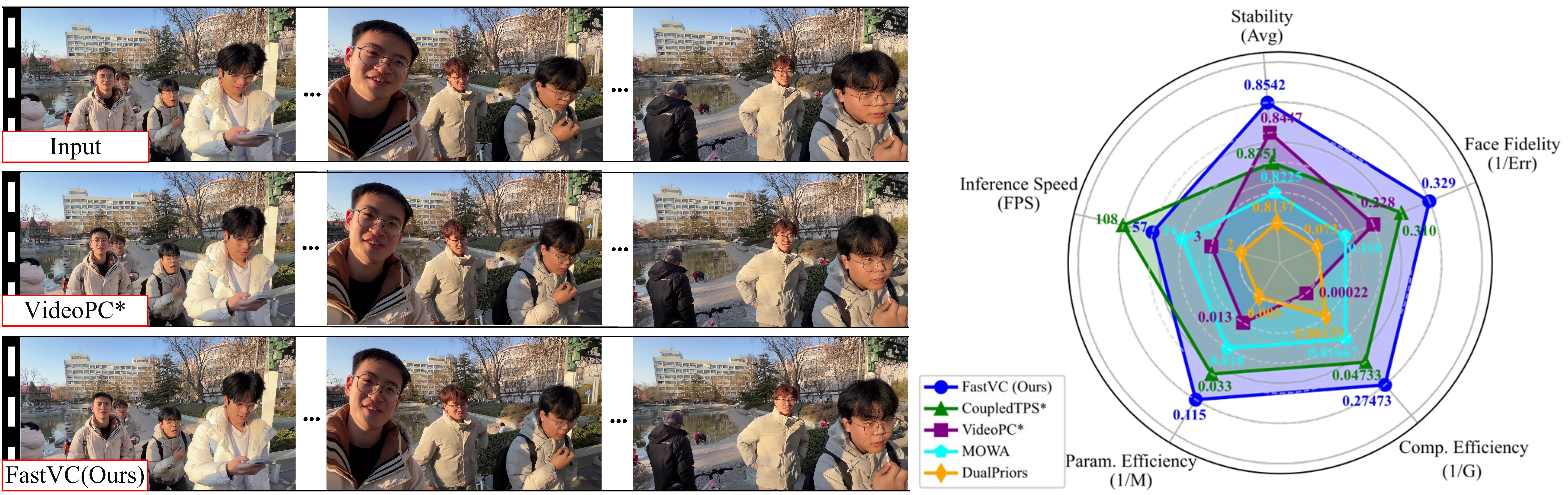}%
      \vspace{2pt}%
      \refstepcounter{figure}%
      \raggedright{\small\textbf{Figure \thefigure.} Comparisons with existing methods on rectification quality, temporal smoothness, and efficiency. Our method achieves real-time full-body correction with temporal stability.}%
      \label{teaser}%
      \vskip 1em plus 2fil%
    }%
  }%
  \newlength\actualheight%
  \settoheight{\actualheight}{\usebox{\titlearea}}%
  \ifdim\actualheight>\titlebox%
    \setlength{\titlebox}{\actualheight}%
  \fi%
  \vbox to \titlebox {%
    \let\footnote\thanks\relax%
    \setcounter{footnote}{0}%
    \def\equalcontrib{%
      \ifnum\value{aaai@eqfn}=0%
        \footnote{These authors contributed equally.}%
        \setcounter{aaai@eqfn}{\value{footnote}}%
      \else%
        \footnotemark[\value{aaai@eqfn}]%
      \fi%
    }%
    \def\corresponding{%
      \ifnum\value{aaai@corrfn}=0%
        \footnote{Corresponding author\ifaaai@corrmulti{s}\fi.}%
        \setcounter{aaai@corrfn}{\value{footnote}}%
      \else%
        \footnotemark[\value{aaai@corrfn}]%
      \fi%
    }%
    \hsize\textwidth%
    \linewidth\hsize%
    \vskip 0.625in minus 0.125in%
    \centering%
    {\LARGE\bf \@title \par}%
    \vskip 0.1in plus 0.5fil minus 0.05in%
    {\Large{\textbf{\aaai@theauthors\ifhmode\\\fi}}}%
    \vskip .2em plus 0.25fil%
    {\normalsize \aaai@affiliations\ifhmode\\\fi}%
    \vskip 0.5em%
    \includegraphics[width=\textwidth,height=0.28\textheight,keepaspectratio]{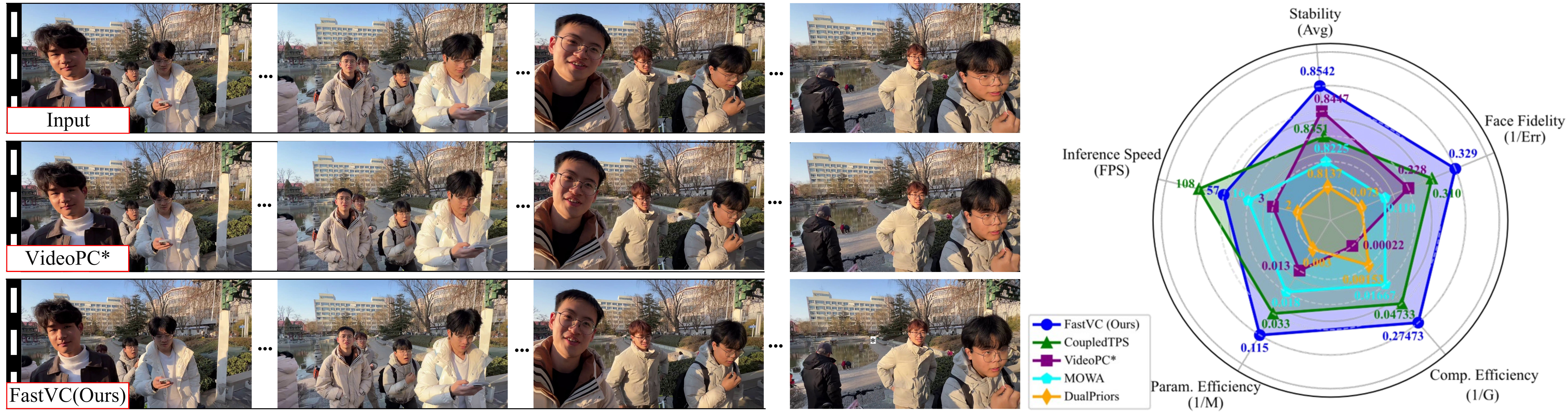}%
    \vspace{2pt}%
    \raggedright{ \small\textbf{Figure 1.} Comparisons with existing methods on rectification quality, temporal smoothness, and efficiency. Our method achieves real-time full-body correction with temporal stability. * indicates
that this model was retrained in the proposed dataset.\vspace{-0.2cm}}%
    \vskip 1em plus 2fil%
  }%
}
\makeatother

\renewcommand{\footnoterule}{%
  \kern -3pt
  \hrule width 2.5cm height 0.4pt
  \kern 2.6pt
}
\begin{document}
\title{Distortion-Aware Spatiotemporal Learning for Real-Time Wide-Angle \\Video Rectification}

\author{
Lang Nie\textsuperscript{1},
Yihan Yang\textsuperscript{1},
Wenbo Nie\textsuperscript{2},
Hongguang Zhu\textsuperscript{3},
Yang Wei\textsuperscript{1},
Chunyu Lin\textsuperscript{2},
Bin Xiao\textsuperscript{1, $\dagger$}
}

\affiliations{
\textsuperscript{1} CQUPT   \hspace{0.5cm}    
\textsuperscript{2} BJTU        \hspace{0.5cm}
\textsuperscript{3} CityU Macau\\
\texttt{\mdseries nielang@cqupt.edu.cn, s250201144@stu.cqupt.edu.cn}
}
\maketitle
\begingroup
\renewcommand{\thefootnote}{}
\footnotetext{$\dagger$ Corresponding author}
\endgroup

\begin{abstract}

While wide-angle lenses provide an expanded field of view (FoV), they exhibit severe nonlinear distortion near image boundaries, unnaturally stretching people in these regions.
Existing methods mostly focus on facial regions, require high computational costs, and lack temporal modeling, resulting in incomplete full-body rectification, impractical real-time deployment, and noticeable inter-frame shakes.
To tackle these issues, we propose \textbf{FastVC}, an efficient full-body video rectification framework driven by distortion-aware spatiotemporal learning. Grounded in lens physics, FastVC introduces a distortion-aware thin-plate spline (DA-TPS) module that nonlinearly remaps control points into a peripheral-dense layout, enabling more accurate geometric fitting without additional computational cost. We further develop a dual-level spatiotemporal learning scheme for causal and stable video rectification. At the model level, neighborhood spatiotemporal attention (NST) aggregates local context and selectively fuses historical features with low overhead. At the optimization level, sparse spatiotemporal supervision applies spatial alignment only to anchor frames, decoupling frame-wise accuracy from temporal stability. Besides, we contribute the first benchmark dataset for wide-angle full-body video correction. Extensive experiments show that FastVC achieves state-of-the-art rectification fidelity and visual stability at 57.0 FPS.
\end{abstract}





\section{Introduction}
Wide-angle lenses are now standard in mobile photography for their expanded FoV. However, this comes at the cost of severe perspective distortion, which unnaturally stretches subjects near frame edges and degrades the quality of both photos and videos. Driven by this, there is a growing demand for preserving natural likeness in wide-angle photography.

Early research formulated portrait correction as a mesh-based energy minimization problem. By designing energy functions like facial correction and line preservation, these methods iteratively solve for an optimal mesh deformation. However, they are computationally expensive and lack robustness, as errors in upstream priors (\textit{e.g.}, face detection, line segment extraction) can easily propagate to the rectified result. Recently, learning-based solutions have emerged as the dominant paradigm for portrait correction, transforming it into a task of end-to-end warping field estimation. By learning complex non-linear mappings from distorted inputs to rectified outputs, these data-driven approaches significantly enhance both inference efficiency and correction precision. 
However, the computational cost of existing state-of-the-art (SoTA) methods still falls short of the real-time requirements for mobile deployment. Beyond this efficiency gap, most models—originally designed for static imagery and facial regions—fail to maintain temporal smoothness across video frames and yield an imbalance between the rectified face and unaltered body regions.  

To overcome these limitations, we propose FastVC, an efficient full-body rectification model for wide-angle videos. A key physical characteristic of wide-angle optics is that perspective distortion is spatially non-uniform, escalating drastically from the optical center toward the periphery. Conventional uniform meshes overlook this distortion pattern, either wasting model capacity on mild center regions or under-fitting severe marginal deformations. To this end, FastVC incorporates a distortion-aware thin-plate spline (DA-TPS) module. By modeling the radial distortion profile, DA-TPS dynamically shifts control points outward to create a peripheral-dense layout. This targeted spatial remapping concentrates the model’s capacity precisely where geometric deformation is most severe, achieving high-fidelity full-body rectification without extra computational overhead.

Besides, to enable causal and stable video rectification, we develop a dual-level spatiotemporal learning scheme spanning model design and optimization. At the model level, we introduce neighborhood spatiotemporal attention (NST), which first encodes neighborhood structures within each frame and then aggregates local features from historical frames using a learnable temporal prior. This avoids costly global interactions while retaining useful spatiotemporal context. At the optimization level, reliable ground truth for real-world wide-angle videos is unavailable, and training must instead rely on temporally inconsistent frame-wise pseudo-labels. We revisit existing dense spatiotemporal constraints that inevitably create a conflict between accurately fitting pseudo-labels and enforcing temporal smoothness.  Then we reformulate it as sparse spatiotemporal supervision with sparse anchor frames, and propagate rectification cues to the other frames throughout the sequential smoothness. It decouples the optimization of frame-wise accuracy from temporal stability, reducing inter-frame jitter and training costs.

As there is no proper dataset readily available for training, we build the first wide-angle video dataset for \textbf{full-body} correction with diverse smartphones and scenes. We annotate it with per-frame pseudo-labels and use it to validate the effectiveness of our sparse spatiotemporal supervision.
Experiments show that FastVC can correct wide-angle videos with high-fidelity visual appearance and temporal stability in real-time. In sum, the contributions center around:
\begin{itemize}
    \item We present FastVC, a pioneering real-time framework that efficiently corrects full-body distortions in wide-angle videos, departing from conventional face-only correction.
    \item FastVC contains a distortion-aware TPS and a dual-level spatiotemporal learning scheme with low overhead. The former fits wide-angle distortion via TPS control points, while the latter leverages neighborhood attention and sparse supervision to enforce spatiotemporal coherence.
    \item We contribute a new video dataset for wide angle full body rectification, serving as a dedicated benchmark to facilitate training and standardized evaluation.

\end{itemize}
\begin{figure*}[t]
\vspace{-0.2cm}
    \centering
    \includegraphics[width=1.0\linewidth]{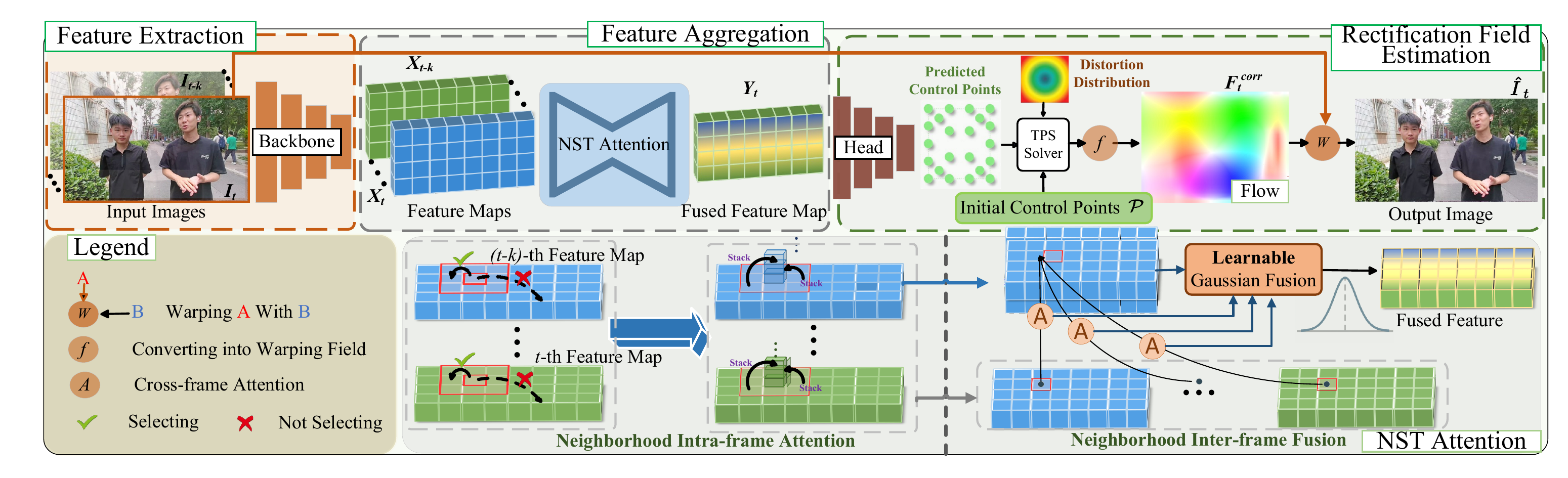} 
    \caption{The overall pipeline of our FastVC network, which consists of feature extraction, feature aggregation, and rectification field estimation. 
    The NST Attention aggregates sequential feature maps into the current frame from spatial to temporal.\vspace{-0.2cm}}
    \label{fig:pipeline}
\end{figure*}

\section{Related Work}
\textbf{Image Portrait Correction:} Wide-angle portrait correction has evolved from traditional geometric optimization to deep learning frameworks. Early optimization-based methods, represented by the works of Carroll et al.~\cite{carroll2009optimizing} and Shih et al.~\cite{shih2019distortion}, established the foundation by constructing mesh-based energy functionals to balance stereographic projection for faces and perspective projection for backgrounds. Recently, MaDCoW~\cite{zhang2025madcow} leveraged the Direct View Condition (DVC) to compute dedicated linear projections for user-specified RoIs and globally fused multi-region constraints. However, these optimization-based approaches heavily rely on line extraction or manual annotations, with some requiring camera intrinsics, while also suffering from high computational costs due to their iterative nature. The advent of deep learning shifted the paradigm toward end-to-end warping field estimation. Tan et al.\cite{tan2021practical} proposed the first cascaded network to decouple background and facial rectification, while Zhu et al.~\cite{zhu2022semi} introduced Transformer-based architectures to capture global semantic dependencies and addressed label scarcity through semi-supervised learning. Subsequent advancements, such as CoupledTPS~\cite{nie2024semi}, enhanced the fitting of non-linear distortions by coupling multiple TPS transformations. More recently, the emergence of MOWA~\cite{liao2025mowa} and UniRect~\cite{qiu2026unirect} marked a new era of "multi-task unification", employing Mamba-based architectures and prompt-guided mechanisms to jointly model portrait correction alongside various other distortion-rectification tasks. 
Despite these successes, image-level methods remain inadequate for maintaining temporal stability in video sequences.

\textbf{Video Portrait Correction:} Extending image correction to video scenarios yields a critical challenge of temporal consistency. Lai et al.~\cite{lai2021correcting} pioneered this field by extending the dual-projection framework with spatiotemporal smoothing constraints, addressing facial trajectory fluctuations and background jumps. However, such optimization-based schemes are too slow for real-time mobile processing. To eliminate the need for laborious video labels, VideoPC~\cite{nie2026beyond} introduced an unsupervised spatiotemporal adaptation framework that leverages pre-trained image correction priors as pseudo-labels to guide video rectification via spatiotemporal adaptation. Despite these efforts, current video-based solutions remain computationally intensive and only focus on facial regions. 

\section{Method} 
Given a sequence of wide-angle video frames $\mathcal{V} = \{I_{1}, I_{2}, \dots, I_N\}$, our goal is to generate the rectified frames $\hat{\mathcal{V}} = \{\hat{I}_{1}, \hat{I}_{2}, \dots, \hat{I}_N\}$ with temporal smoothness. 
\vspace{-0.1cm}
\subsection{Overview}
\label{sec:overview}

To enable causal real-time processing, FastVC avoids high-latency full-video inputs by adopting a sliding-window mechanism. Specifically, given $k$ historical frames $\{I_{t-k}, \dots, I_{t-1}\}$ and the current frame $I_t$, the model predicts the rectified current frame $\hat{I}_t$ in an online manner.

As illustrated in Fig.~\ref{fig:pipeline}, the current wide-angle frame and historical frames are first processed by a shared backbone to produce a
sequence of latent feature maps $(X_{t-k}, \dots, X_t)$. To achieve spatiotemporal feature aggregation at the model level, these maps enter the neighborhood spatiotemporal (NST) attention module, a lightweight aggregator that captures intra-frame local structures and inter-frame temporal dependencies within a confined neighborhood to yield a context-enriched representation $Y_t$. Then $Y_t$ is passed through a regression head to estimate a distortion-aware TPS transformation. This transformation can be converted as a pixel-wise warping field ($F_t^{corr}$), enabling the rectified frame $\hat{I}_t$ to be rendered via  resampling $(W)$.

By sliding the window across $Z$ consecutive steps, the model generates a sequence of rectified outputs $\{\hat{I}_t\}_{t=k+1}^{k+Z}$. 
At the optimization level, we introduce a sparse spatiotemporal supervision strategy (Fig.~\ref{fig:sparse_supervision}) to effectively relieves the conflict of spatial accuracy and temporal smoothness with sparse anchor frames. 
\subsection{Distortion-Aware TPS Transformation}
\begin{figure}[h]
    \centering
    \includegraphics[width=0.48\textwidth]{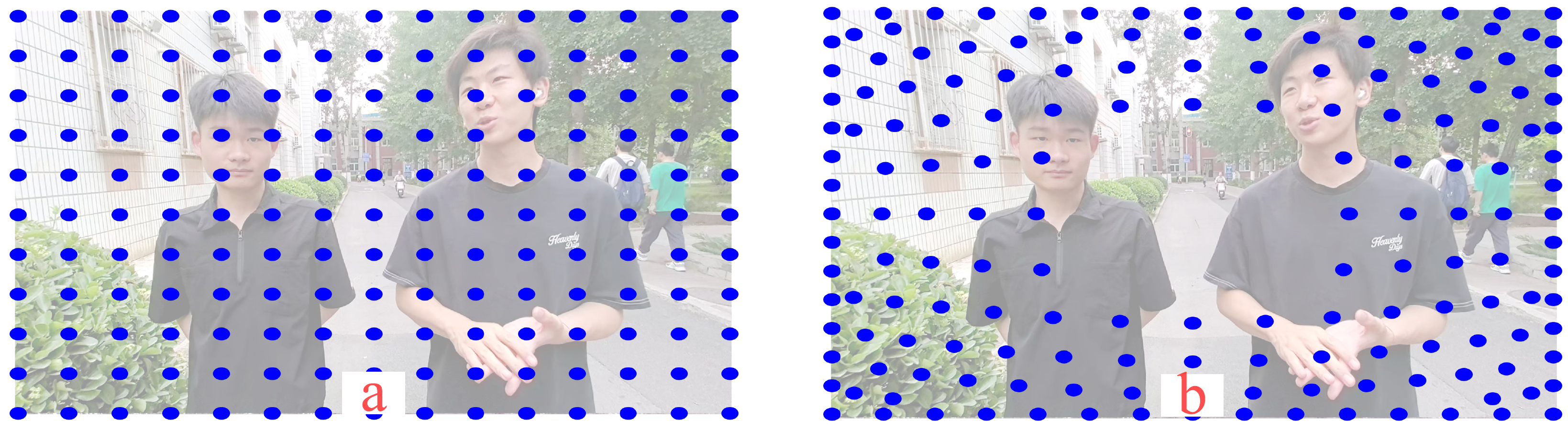}
    \caption{(a) Uniform layout. (b) Distortion-aware layout.\vspace{-0.3cm}}
    \label{fig:tps_points}
\end{figure}

Given the context-enriched feature $Y_t$ from the NST module, we employ a lightweight regression head to predict dynamic control point offsets. Existing TPS-based methods~\cite{nie2024semi,nie2023parallax} typically position control points on a uniform layout (Fig.~\ref{fig:tps_points}a). However, optical distortion in wide-angle lenses follows a non-linear radial profile, which is minimal at the center but becomes severe toward the periphery. Uniform layout inherently wastes capacity on the near-rectilinear center while undersampling heavily distorted boundaries.

To resolve this representation bottleneck, we propose a distortion-aware, non-equidistant control point layout (DA-TPS) that redistributes modeling capacity to align with physical lens priors. Given a uniform coordinate with normalized radial distance $r \in [0, 1]$ relative to the image center, we remap $r$ via a non-linear concave function:
\begin{equation}
\tilde{r} = g(r) = \frac{\log(1 + \alpha \cdot r)}{\log(1 + \alpha)},
\end{equation}
where $\alpha$ controls the radial density gradient. Since $g(r)$ is strictly monotonically increasing and concave ($g''(r) < 0$), it enforces a sparse sampling near the stable center and dense sampling toward the boundaries.

After converting $\tilde{r}$ back to Cartesian coordinates, we perform spatial layout optimization. Specifically, redundant points in the flat center are pruned to eliminate over-parameterization, while equidistant anchor points are appended along the perimeter as topological constraints to safeguard rectangular boundary integrity. As illustrated in Fig.~\ref{fig:tps_points}b, this yields a deterministic, sparse set of $N=172$ active control points $\mathcal{P}$. This layout concentrates the transformation capacity precisely where geometric distortion is most severe.

Given the refined point set $\mathcal{P}$ and predicted displacement vectors, we formulate the TPS interpolation $T: \mathbb{R}^2 \to \mathbb{R}^2$. The pixel-wise warping field $F_t^{\text{corr}}$ for any target coordinate $p$ is derived as:
\begin{equation}
F_t^{\text{corr}}(p) = T(p) - p = C + A(p) + \sum_{p_i \in \mathcal{P}} w_i \cdot U(\| p - p_i \|) - p,
\end{equation}
where $C, A$, and $w_i$ are standard TPS coefficients solved via closed-form linear systems~\cite{kent1994link}, and $U(z) = z^2 \log(z)$ denotes the radial basis function.

By reallocating control points rather than adding points, DA-TPS substantially enhances geometric representation. Spatially, concentrating capacity at distorted peripheries introduces a physical inductive bias for high-fidelity restoration. Temporally, dense boundary points form a smooth spatial manifold that damps grid fluctuations, inherently favoring inter-frame consistency. See supplementary material for full algorithmic steps, derivations, and ablation on $N$.

\subsection{Dual-Level Spatiotemporal Learning Scheme}
\label{sec:dual_level}
 To enable causal and stable video rectification, we propose a dual-level learning scheme. At the model level, it performs local spatiotemporal feature aggregation; at the optimization level, it applies sparse supervision to decouple spatial and temporal constraints.

\subsubsection{Neighborhood Spatiotemporal Attention:}
\label{sec:nst}
At the model level, the Neighborhood Spatiotemporal Attention (NST) essentially serves as a feature fusion mechanism that, conditioned on the current frame ($X_{t}$), selectively incorporates relevant information from historical frames ($X_{t-k}, \dots, X_{t-1}$) to produce a spatiotemporally enhanced representation ($Y_{t}$). To achieve high-fidelity video rectification with minimal overhead, this is realized via a spatial-first, temporal-next hierarchical aggregation, which strictly restricts attention operations to a local spatiotemporal neighborhood. As depicted in Fig.~\ref{fig:pipeline}, it operates through two sequential steps:

\textit{Neighborhood Intra-Frame Attention}: Denoting $X$ as an arbitrary feature map from $\{X_{t-k}, \dots, X_{t}\}$, it is first partitioned into patches and augmented with positional embeddings to obtain $\mathbf{X_p} \in \mathbb{R}^{M \times C}$, where $M$ and $C$ denote the sequence length and embedding dimension. We then project $\mathbf{X_p}$ into query ($Q$), key ($K$), and value ($V$) embeddings. To implement the localized attention mechanism, we define a neighborhood window $\Omega(i)$ for each patch $i$ based on its spatial coordinates:
\begin{equation}
\text{Attn}(Q^i, K, V) = \sum_{j \in \Omega(i)} \text{Softmax} \left( \frac{Q^i (K^j)^T}{\sqrt{d}} \right) V^j,
\end{equation}
where $\Omega(i)$ covers a local $w \times w$ region centered at patch $i$, and $\sqrt{d}$ represents the scaling factor. This localized restriction ensures that each query only interacts with its spatially adjacent patches, which reduces the computational complexity from $O(M^2)$ to $O(M \cdot w^2)$.

\textit{Neighborhood Inter-Frame Fusion}: After carrying out neighborhood intra-frame attention for every input frame, we enrich each patch $i$ by concatenating its features with those of its spatial neighbors within $\Omega(i)$. This concatenated representation is projected back to dimension $C$, explicitly assembling local context into each patch to facilitate robust cross-frame matching. These feature maps are then projected into $Q_{t-\tau}, K_{t-\tau}, V_{t-\tau}$ embeddings ($\tau \in \{0, 1, \dots, k\}$). To aggregate temporal context under the guidance of the current frame, we treat the current frame's query as the reference and attend to corresponding keys/values from historical frames at the same spatial location. Formally, the temporal fusion result $Y^i_t$ for patch $i$ at frame $t$ is formulated as:
\begin{equation}
Y^i_t = \sum_{\tau=0}^{k} \text{Softmax} \left( \frac{Q^i_t (K^i_{t-\tau})^T}{\sqrt{d}} \cdot G(\tau) \right) V^i_{t-\tau}.
\label{eq:temporalfusion}
\end{equation}
Considering the varying importance of frames at different temporal distances, we incorporate a Gaussian weight prior $G(\tau)$ in Eq.~\ref{eq:temporalfusion}. To adapt this prior to real video samples, its variance $\sigma$ is set as a learnable parameter determining how quickly temporal influence diminishes:
\begin{equation}
G(\tau) = \exp\left( -\frac{\tau^2}{2\sigma^2} \right).
\end{equation}
This spatial-to-temporal factorization effectively improves correction accuracy through aggregating spatiotemporal features at the feature level while maintaining a lightweight computational overhead for real-time video applications.
\begin{figure}[h]
    \centering
    \includegraphics[width=1.0\linewidth]{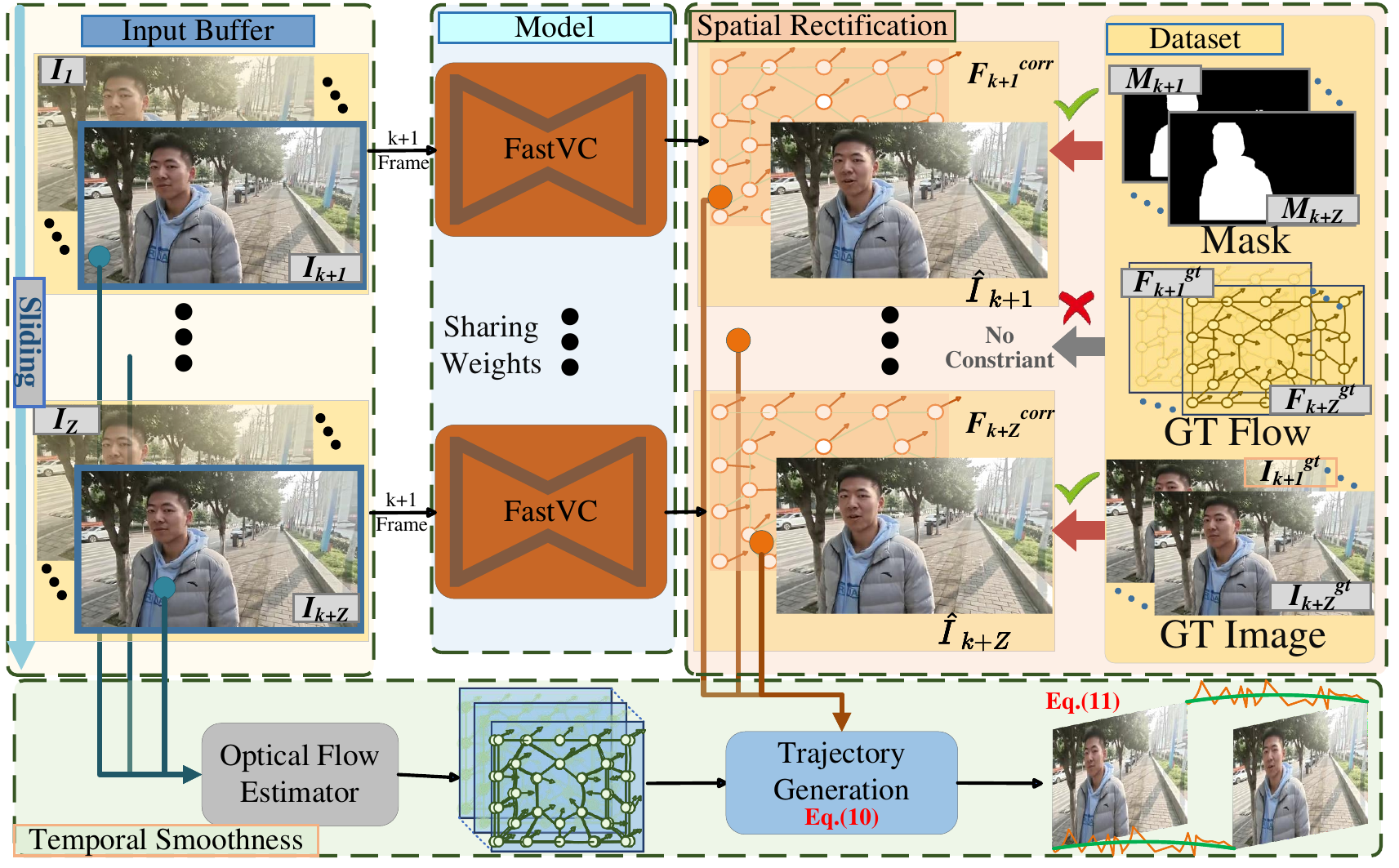} 
    \caption{The sparse spatiotemporal supervision strategy.\vspace{-0.4cm}}
    \label{fig:sparse_supervision}
\end{figure}

\begin{figure*}
    \centering
    \includegraphics[width=1.0\linewidth]{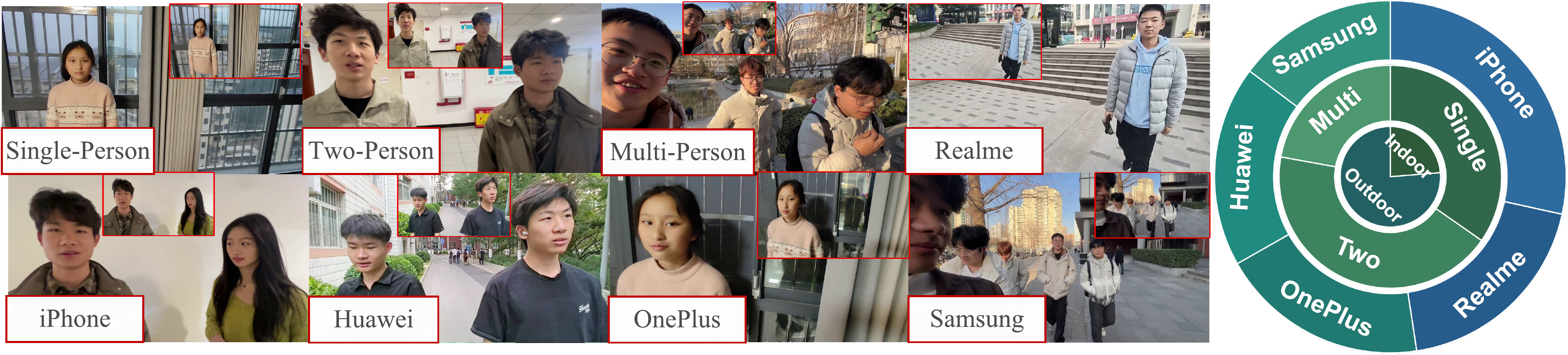} 
    \caption{Overview of the proposed wide-angle full-body video dataset. Left: Representative samples across diverse scenes, subjects, and brands, with insets showing the corresponding ground-truth rectified frames. Right: Hierarchical distribution of the dataset, categorized by scene, number of subjects and smartphone brand.\vspace{-0.2cm}}
    \label{fig:dataset}
\end{figure*}
\subsubsection{Sparse Spatiotemporal Supervision:}
\label{sec:sparse_sup}
At the optimization level, we address the intrinsic trade-off between per-frame geometric accuracy and inter-frame motion stability. An intuitive paradigm for video rectification is dense supervision, applying spatial losses and temporal constraints simultaneously to all $Z$ frames in a sliding window:
\begin{equation}
\mathcal{L}_{\text{dense}} = \sum_{t=k+1}^{k+Z} \mathcal{L}_{\text{spatial}}(t) + \lambda \mathcal{L}_{\text{temp}}.
\label{eq:dense_sup}
\end{equation}
However, we argue that such dense constraint is inherently sub-optimal. Because pseudo-ground-truth labels are synthesized independently per frame~\cite{shih2019distortion}, they inevitably contain subtle spatial misalignment noise. Overfitting these independent spatial targets on every frame (\textit{i.e.}, Eq.~\ref{eq:dense_sup}) forces the network to memorize per-frame artifacts, transferring spatial noise into temporal flickering while incurring heavy training overhead.

\textit{Sparse Supervision Paradigm}: To break this dilemma, we propose a sparse supervision mechanism that decouples spatial fidelity from temporal smoothness. As illustrated in Fig.~\ref{fig:sparse_supervision}, explicit spatial supervision is restricted exclusively to sparse boundary keyframes, while intermediate frames are optimized solely via temporal motion regularization:
\begin{equation}
\mathcal{L}_{\text{total}} = \sum_{t \in \{k+1, k+Z\}} \mathcal{L}_{\text{spatial}}(t) + \lambda \mathcal{L}_{\text{temp}},
\label{eq:sparse_sup}
\end{equation}
where $k+1$ and $k+Z$ denote the anchor frame indices. By liberating intermediate frames from fitting noisy pseudo-labels, $\mathcal{L}_{\text{temp}}$ guides the network to dynamically infer a minimal-energy, trajectory-smooth transition bridging the anchor keyframes.

\textit{Temporal Smoothness Term}: Specifically, the temporal motion smoothness term $\mathcal{L}_{\text{temp}}$ regulates inter-frame coherence by penalizing trajectory abruptness in predicted correction fields. Let $F_{t\to t+1}^{\text{inter}}$ denote the inter-frame optical flow; the spatial position increment $\Delta_{t\to t+1}$ of the predicted flow $F_t^{\text{corr}}$ between adjacent frames is derived as:
\begin{equation}
\Delta_{t\to t+1} = \mathcal{W}(F_{t\to t+1}^{\text{inter}}, F^{\text{corr}}_t) + F^{\text{corr}}_t - F^{\text{corr}}_{t+1},
\end{equation}
where $\mathcal{W}(\cdot)$ is the backward warping operator. Rather than calculating long-range paths, we encourage consecutive frame-wise displacement increments across the sequence to evolve smoothly:
\begin{equation}
\mathcal{L}_{\text{temp}} = \frac{1}{Z-2}\sum_{t=1}^{Z-2} \| \Delta_{t\to t+1} - \Delta_{t+1\to t+2} \|_1.
\label{eq:temporal}
\end{equation}

\textit{Spatial Rectification Term}: For the sparse anchor keyframes $t \in \{k+1, k+Z\}$, the spatial loss $\mathcal{L}_{\text{spatial}}(t)$ ensures precise structural alignment. To focus optimization on human body regions, a binary body mask $M_t$ is integrated into a multi-scale spatial loss formulation:
\begin{equation}
\mathcal{L}_{\text{spatial}}(t) = \lambda_1\mathcal{L}_{\text{pixel}}(t) + \lambda_2\mathcal{L}_{\text{gradient}}(t) + \lambda_3\mathcal{L}_{\text{feature}}(t),
\label{eq:spatial}
\end{equation}
where $\lambda_1, \lambda_2, \lambda_3$ balance the individual components. The pixel-level flow loss $\mathcal{L}_{\text{pixel}}(t)$ minimizes coordinate mapping errors:
\begin{equation}
\mathcal{L}_{\text{pixel}}(t) = \| (M_t+1) \odot (F^{\text{corr}}_t - F^{\text{gt}}_t)\|_1.
\end{equation}
The gradient-level flow loss $\mathcal{L}_{\text{gradient}}(t)$ preserves geometric edge structures along both horizontal and vertical axes via Sobel operators $\nabla_x$ and $\nabla_y$:
\begin{equation}
\begin{aligned}
\mathcal{L}_{\text{gradient}}(t) = &\| (M_t+1) \odot (\nabla_x F^{\text{corr}}_t - \nabla_x F^{\text{gt}}_t)\|_1 \\
+ &\| (M_t+1) \odot (\nabla_y F^{\text{corr}}_t - \nabla_y F^{\text{gt}}_t)\|_1.
\end{aligned}
\end{equation}
Finally, the feature-level appearance loss $\mathcal{L}_{\text{feature}}(t)$ maintains semantic structural consistency through a pre-trained VGG feature extractor $\phi(\cdot)$:
\begin{equation}
\mathcal{L}_{\text{feature}}(t) = \| (M_t+1)\odot(\phi(\hat{I}_t) - \phi(I_t^{\text{gt}})) \|_2.
\end{equation}

\vspace{0.15cm}
\section{Dataset and Metrics}
\subsection{Dataset Construction} 
Due to the lack of dedicated benchmarks for wide-angle full-body video portrait rectification, we construct a comprehensive dataset containing 32 1080P clips at 30fps, totaling 17,029 frames with durations of 9--35s. 
Captured with five mainstream smartphone brands (\textit{e.g.}, iPhone and Samsung) across diverse real-world scenarios, the dataset is split into 22 training sequences (11,814 frames) and 10 test sequences (5,215 frames).
To generate full-body correction labels, we adopt the image-level rectification~\cite{shih2019distortion} by expanding the RoI from faces to full bodies, which uses YOLO11m-seg instance masks as spatial weight guidance and metric references. Each annotation is a quadruple: \text{wide-angle image}, \text{rectified GT}, \text{flow GT}, and \text{full-body mask}.

As summarized in Table~\ref{tab:dataset_comparison}, existing datasets are restricted to facial rectification. Besides, Tan et al.~\cite{tan2021practical} provide only 5.3k static images, whereas VideoPC~\cite{nie2026beyond} uses merely 7.1k of its 44k unlabeled frames for temporal fine-tuning.
In contrast, our wide-angle full-body video rectification dataset contains 17k frames with image-level full-body labels.
Together, its geometric diversity, temporal coverage, and frame-level annotations support robust model training with sparse spatiotemporal supervision while reducing reliance on temporally inconsistent dense labels.
\begin{table}[t]
\centering
\small
\setlength{\tabcolsep}{4pt} 
\resizebox{\linewidth}{!}{
\begin{tabular}{lcccc}
\toprule
\textbf{Dataset} & \textbf{Type} & \textbf{Target} & \textbf{GT} & \textbf{Training/Total Frames} \\ \midrule
Tan et al. & Image & Face & Image Label & ~5.2k/5.3k \\
VideoPC  & Video & Face & No Label & 7.1k/44k \\
\textbf{Ours} & \textbf{Video} & \textbf{Full-Body} & Image Label & \textbf{12k}/17k \\ \bottomrule
\end{tabular}}
\caption{Comparison of wide-angle portrait datasets.\vspace{-0.3cm}}
\label{tab:dataset_comparison}
\vspace{-0.3cm}
\end{table}


\begin{table*}[t]
\centering
\small
\setlength{\tabcolsep}{4pt}
\scalebox{1.1}{
\begin{tabular}{lccccccccccc}
\toprule
\multirow{2}{*}{Method} & \multirow{2}{*}{Reference}   & \multicolumn{3}{c}{Face AR (\%)$\downarrow$} & \multicolumn{3}{c}{Body AR (\%)$\downarrow$} & \multicolumn{3}{c}{Stability Score $\uparrow$} \\
 &  & \textbf{Avg. Diff. }  & \textbf{$\bar{e}_{5}$}& \textbf{$\bar{e}_{10}$} &\textbf{Avg. Diff. } & \textbf{$\bar{e}_{5}$}& \textbf{$\bar{e}_{10}$} &Avg. &Trans. & Rot.\\
\midrule
DualPriors  & ECCV2024 &  13.73 & 84.59 & 67.39 & 6.18 & 47.74 & 18.99 & 0.8137 & 0.7878 &0.8396 \\
CoupledTPS    & TPAMI2024  &9.07&  75.30&  41.29 & 5.59 &  40.57& 16.97&  0.8352&  0.8058&0.8646  \\
CoupledTPS*  & TPAMI2024  &  3.23& 21.06 &4.41  & 2.42 & 13.87 & 2.76 & 0.8351 & 0.7860 & \textbf{0.8842} \\
MOWA  & TPAMI2025& 9.08 & 74.92 & 37.59 & 5.87 & 42.93 & 18.66 & 0.8225 & 0.7908 & 0.8542 \\
VideoPC  & AAAI2026  & 9.66 & 75.15 & 48.84 & 5.56 & 50.60 & 11.06 & 0.7996 & 0.7423 & 0.8569 \\
VideoPC*  & AAAI2026  &4.38  & 38.93 & 3.78 & 3.16 & 18.91 &4.67  & 0.8447 &  0.8080& 0.8814 \\  
\textbf{FastVC (Ours)}  & -&\textbf{3.04} &\textbf{17.36}&\textbf{2.63}&\textbf{2.35}&\textbf{13.09}&\textbf{2.67}&\textbf{0.8542}&\textbf{0.8265}&0.8818
 \\
\bottomrule
\end{tabular}%
}
\caption{Quantitative comparison with state-of-the-art methods on spatial rectification quality and temporal stability. 
* indicates that this model was retrained in the proposed dataset.
\vspace{-0.2cm}}
\label{comparison_results}
\end{table*}
\subsection{Evaluation Metric}
\label{sec:metrics}
Existing rectification studies predominantly focus on images and are commonly evaluated using image-level metrics such as $\text{LineAcc}$ and $\text{ShapeAcc}$~\cite{tan2021practical}. However, these metrics are ill-suited to our wide-angle full-body video setting. First, they assess frames independently and therefore cannot measure temporal consistency. Second, they require spatially aligned static references and are sensitive to small positional deviations introduced by temporal smoothing, even when the outputs remain perceptually stable. Moreover, $\text{ShapeAcc}$ considers only facial landmarks and cannot adequately evaluate full-body deformation.

Given the lack of dedicated metrics for wide-angle video rectification, we introduce the \textbf{average difference of relative aspect ratio (Avg. Diff.)} for both face and body regions:
\begin{equation}
    \text{Avg. Diff.} = \frac{1}{\sum_{i=1}^{N} P_i} \sum_{i=1}^{N} \sum_{j=1}^{P_i} \frac{|R_{\text{pred}}^{(i,j)} - R_{\text{gt}}^{(i,j)}|}{R_{\text{gt}}^{(i,j)}} \times 100\%,
\end{equation}
where $N$ is the number of frames, $P_i$ is the person count in frame $i$, and $R_{\text{pred}}^{(i,j)}$ and $R_{\text{gt}}^{(i,j)}$ denote the aspect ratios (width/height) of the $j$-th person in the predicted and ground-truth frames, respectively. By measuring the aspect ratio variations of these RoIs instead of absolute coordinates, this metric remains invariant to minor spatial shifts. We also report the percentage of frames where the error exceeds 5\% and 10\% ($\bar{e}_{5}$, $\bar{e}_{10}$) to evaluate overall robustness.

Furthermore, to evaluate the stability of the rectified video, we employ the commonly used video stability metrics \textbf{Stability Score}~\cite{choi2020deep} in the field of video stabilization~\cite{liu2013bundled}, which calculates the inter-frame smoothness of the rectified sequence. These metrics collectively ensure that the model not only restores natural human proportions but also maintains the fluid visual experience required for high-quality video processing applications.

\vspace{-0.1cm}
\section{Experiment}

\vspace{-0.cm}
\subsection{Implementation Details}
\label{sec:51}
We implement FastVC in PyTorch and conduct training on a single NVIDIA RTX 5090 GPU for 120 epochs with a batch size of 7. Video frames are resized to $512 \times 384$. We set the TPS control points to $N=172$ ($\alpha=1$) and adopt the Adam optimizer with an initial learning rate of $1\times 10^{-4}$. Loss weights are set as $\lambda_1 = 2 \times 10^{-4}$, $\lambda_2 = 1$, $\lambda_3 = 10$, and $\lambda = 3$. Optical flow for temporal regularization is estimated via a pre-trained RAFT. Full hyper-parameter configurations are provided in the supplementary material.

\vspace{-0.1cm}
\subsection{Comparison Result}
\vspace{-0.1cm}
\label{sec:52}
\begin{figure*}[htbp]  
    \centering  
    \includegraphics[width=0.93\textwidth]{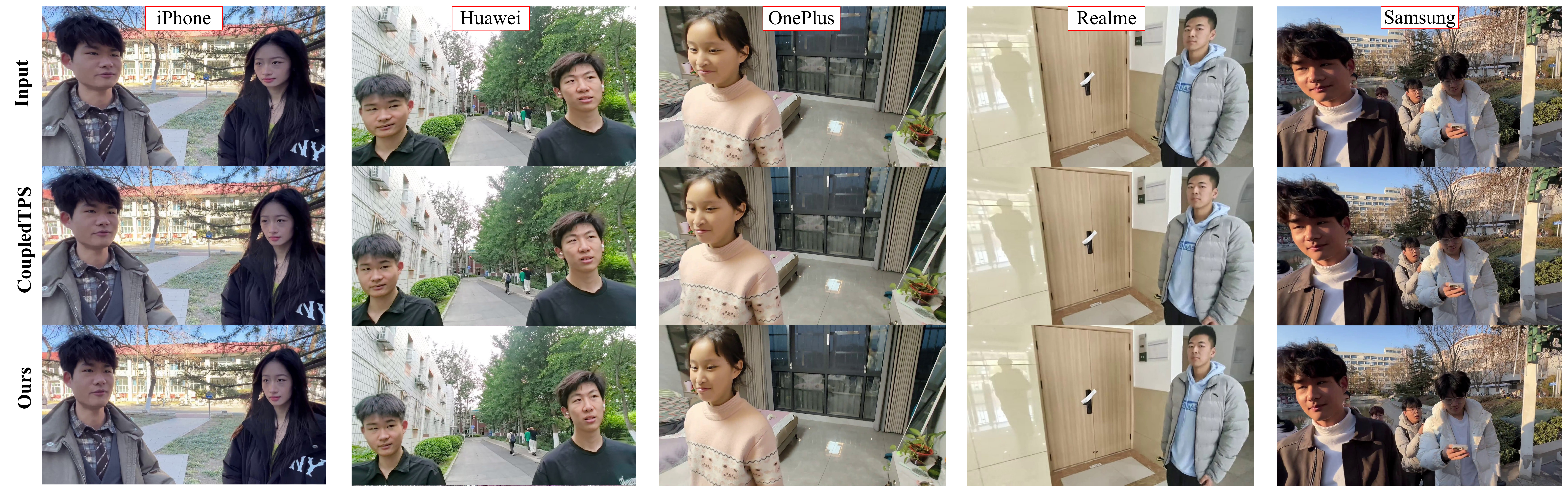}
    \caption{Qualitative comparison across various mobile device brands, demonstrating our superior full-body rectification quality.\vspace{-0.1cm}}  
    \label{fig:brand}  
\end{figure*}

\vspace{0.1cm}
\noindent\textbf{Rectification Performance: }
As quantitatively summarized in Table~\ref{comparison_results}, FastVC consistently outperforms both state-of-the-art image-based and video-based models. Specifically, FastVC achieves the lowest distortion errors on both face (3.04\% Avg. Diff.) and body (2.35\% Avg. Diff.) regions. The notable reduction in extreme error $\bar{e}_{10}$ (from 4.41\% to 2.63\% on faces) demonstrates its superior capability in mitigating severe wide-angle perspective distortions. Qualitative results across diverse smartphone brands in Fig.~\ref{fig:brand} further highlight our simultaneous face-body restoration fidelity.

In terms of video consistency, FastVC achieves a leading translation stability score (0.8265), effectively preventing the high-frequency temporal flickering commonly suffered by image-based baselines (\textit{e.g.}, MOWA~\cite{liao2025mowa}, CoupledTPS~\cite{nie2024semi}).
\vspace{-0.3cm}
\begin{table}[h] 
\centering
\resizebox{\linewidth}{!}{%
\begin{tabular}{lccc}
\toprule
\textbf{Method} & \textbf{Correction} $\uparrow$ & \textbf{Stability} $\uparrow$ & \textbf{Background Fidelity} $\uparrow$ \\ \midrule
CoupledTPS* & 10.59\% & 1.76\% & 2.35\% \\
VideoPC*    & 2.94\% & 43.53\% & 35.88\% \\
\textbf{FastVC (Ours)} & \textbf{86.47\%} & \textbf{54.71\%} & \textbf{61.77\%} \\ \bottomrule
\end{tabular}%
}
\caption{Quantitative comparison of the user study. }
\label{tab:user_study}
\vspace{-0.3cm}
\end{table}

\noindent\textbf{User Study:} 
To assess human visual preference, we conduct a pairwise blind user study with 20 participants across 10 testing videos (5 brands), each segmented into one-second clips(170 clips in total). As shown in Table~\ref{tab:user_study}, FastVC achieves overwhelming dominance in \textit{Correction Quality} (86.47\%), \textit{Stability} (54.71\%), and \textit{Background Fidelity} (61.77\%), reinforcing our quantitative findings.
\begin{table}[htbp]
\vspace{-0.3cm}
  \centering
  \scalebox{0.85}{
  \begin{tabular}{ccccc}
    \toprule
    Type& Method & Params (M) & GFLOPs  & FPS \\
    \midrule
    \makecell{Image Model} & \makecell{DualPriors\\MOWA\\CoupledTPS} & \makecell{400.0\\55.6\\30.6} & \makecell{654.96\\59.99\\21.13} & \makecell{2.2\\16.4\\ \textbf{107.9}}\\

    \midrule
    \makecell{Video Model} &\makecell{VideoPC\\FastVC (Ours)}  &\makecell{80.0\\\textbf{8.7}}  &\makecell{4515.16\\\textbf{3.64}}  &\makecell{2.88\\57.0} \\
    \bottomrule
  \end{tabular}
  }
  \caption{Efficiency comparison on a single RTX 5090 GPU.}
  \label{efficiency}
\end{table}
\vspace{-0.3cm}

\noindent\textbf{Efficiency Comparison:}
\begin{table*}[htbp]
\centering
\small
\setlength{\tabcolsep}{4pt}
\scalebox{1.0}
{
\begin{tabular}{ccccccccccc}
\toprule

Supervision & Output Frames & \multicolumn{3}{c}{Face AR (\%)$\downarrow$} & \multicolumn{3}{c}{Body AR (\%)$\downarrow$} & \multicolumn{3}{c}{Stability Score $\uparrow$} \\
 Mode & During Training ($Z$)& \textbf{Avg. Diff. }  & \textbf{$\bar{e}_{5}$}& \textbf{$\bar{e}_{10}$} &\textbf{Avg. Diff. } & \textbf{$\bar{e}_{5}$}& \textbf{$\bar{e}_{10}$} &Avg. &Trans. & Rot.\\
\midrule
Spatial Term (Eq.~\ref{eq:spatial})& 3 & 2.98 &19.14 &1.29 & 2.36 &13.69& 2.53 & 0.8483& 0.8189&0.8777  \\
Temporal Term (Eq.~\ref{eq:temporal})& 3 & \multicolumn{9}{c}{Training Collapse}\\
Dense Spatiotemporal (Eq.~\ref{eq:dense_sup})& 3 & 3.84 & 22.26&7.52 & 2.58 &16.25& 3.05 &\textbf{0.8555}&0.8226 &\textbf{ 0.8883} \\
Sparse Spatiotemporal (Eq.~\ref{eq:sparse_sup})& 3  &\textbf{3.04} &\textbf{17.36}&2.63&\textbf{2.35}&\textbf{13.09}&2.67&0.8542&\textbf{0.8265}&0.8818
 \\
\midrule
Sparse Spatiotemporal (Eq.~\ref{eq:sparse_sup})& 5 & 3.54&22.41 & 3.20 & 2.44 &13.42& \textbf{2.51} &0.8542 & 0.8224 & 0.8859 \\
Sparse Spatiotemporal (Eq.~\ref{eq:sparse_sup})& 7  & 3.39 &21.39& \textbf{2.47} & 2.38 &13.50& 2.59 &0.8375 & 0.7912 & 0.8838 \\
\bottomrule
\end{tabular}%
}
\caption{Ablation study of training strategies, including supervision mode and output frames during training. \vspace{-0.2cm}
}
\label{ablation}
\end{table*}
Table~\ref{efficiency} evaluates computational efficiency on a single RTX 5090 GPU. Owing to our lightweight design, FastVC requires only 8.7M parameters and 3.64 GFLOPs, substantially reducing computational complexity compared to heavy counterparts like VideoPC (4515.16 GFLOPs). Although CoupledTPS achieves higher FPS via vanilla hardware-optimized convolutions, FastVC comfortably operates at 57.0 FPS, well above the real-time requirement ($>30$ FPS) while offering superior rectification accuracy and temporal stability.

\begin{figure}
    \centering
    \includegraphics[width=1.0\linewidth]{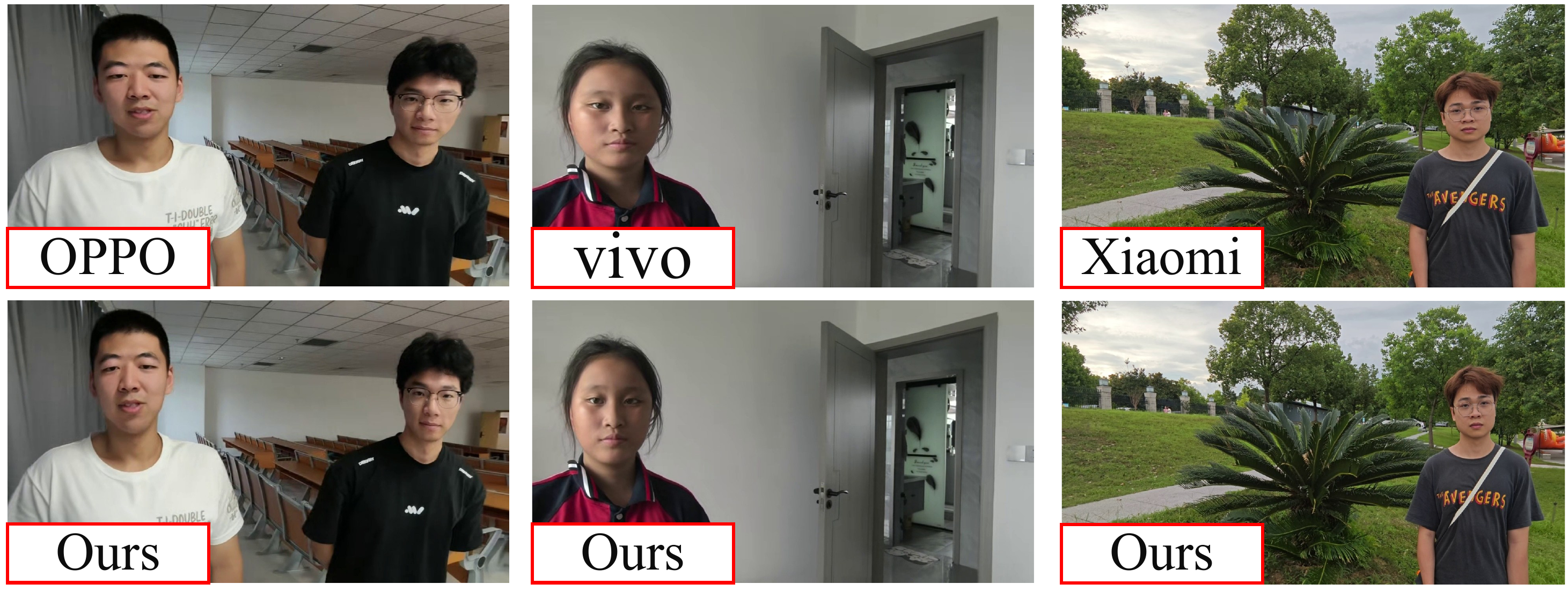}
    \caption{Qualitative results on unseen camera devices.}
    \label{fig:generalization}
    \vspace{-0.6cm}
\end{figure}

\noindent\textbf{Cross-Device Generalization:} To evaluate generalizability, we test FastVC with three unseen brands (OPPO, vivo, and Xiaomi) absent from our dataset. As shown in Fig.~\ref{fig:generalization}, without fine-tuning, FastVC consistently restores natural facial and body proportions, as well as maintains background straight lines. This substantiates that the network captures intrinsic distortion priors, not sensor‑specific overfitting, and we refer readers to the supplementary material for full videos.

\vspace{-0.2cm}
\subsection{Ablation Study}
\label{sec:53}


\vspace{-0.cm}
\noindent\textbf{Training Strategy: }
Table~\ref{ablation} evaluates different supervision modes and sequence lengths ($Z$). 
First, training with the temporal term alone leads to severe training collapse due to the absence of geometric spatial anchor constraints. 
Second, while dense spatiotemporal supervision achieves high raw stability, it exhibits severely degraded spatial fidelity (3.84\% Face AR and 7.52\% $\bar{e}_{10}$) because forcing every frame to overfit independent noisy pseudo-labels introduces spatial-temporal interference. 
In contrast, our sparse supervision restricts spatial constraints to anchor keyframes, relieving intermediate frames from label noise while maintaining comparable stability (0.8542 vs. 0.8555). 
Lastly, setting $Z=3$ yields the best trade-off, whereas larger window sizes ($Z=5, 7$) introduce error accumulation over long temporal spans.

\begin{table}[htbp]
\centering
\resizebox{\linewidth}{!}{%
\begin{tabular}{lccccc}
\toprule

\makecell{Input \\Frames} & \makecell{NST} &\makecell{DA-TPS} &\makecell{Face AR (\%)$\downarrow$\\Avg. Diff.} &\makecell{Body AR (\%)$\downarrow$\\Avg. Diff.} &\makecell{Stability Score $\uparrow$\\Avg.} \\

\midrule
\makecell{1} &  && 3.61 &  2.45 &   0.8272\\
\makecell{1} & &\checkmark & \textbf{2.74} & 2.41 & 0.8426 \\
\midrule
\makecell{3}&  && 3.20 & 2.67  & 0.8345\\
\makecell{3} &  &\checkmark& 3.30  & 2.63 &   0.8529 \\
\makecell{3} & \checkmark && 3.17 &  \textbf{2.33}  & 0.8378 \\ 
\makecell{3} & \checkmark &\checkmark& 3.04  & 2.35  & \textbf{0.8542} \\ 
\bottomrule
\end{tabular}%
}
\caption{Ablation study on model architecture. Default grid uses 180 uniform control points when DA-TPS is disabled. Full metrics are provided in the supplementary material.\vspace*{-0.2cm}}
\label{single_vs_multi}
\vspace*{-0.1cm}
\end{table}
\noindent\textbf{Model Architecture: }
We evaluate our key architectural components in Table~\ref{single_vs_multi}. Under the single-frame setup (Rows 1--2), integrating DA-TPS consistently boosts all metrics, notably reducing Face AR from 3.61\% to 2.74\%. This confirms that aligning control point density with physical lens priors enhances geometric expressiveness without extra parameters.

Expanding to 3 frames, a fundamental tension arises between spatial accuracy and temporal consistency. Naive temporal concatenation introduces feature misalignment that degrades geometric accuracy, worsening Body AR from 2.45\% to 2.67\%. 
Equipping the network with NST (Row 5) mitigates this bottleneck. NST leverages its spatiotemporal aggregation capability to boost spatial performance, improving Body AR to 2.33\% and Face AR to 3.17\% over Row 3 while maintaining a 0.8378 temporal stability. Fully integrating NST and DA-TPS (Row 6) yields the optimal overall balance, achieving the lowest Face AR (3.04\%) and highest Stability Score (0.8542). This demonstrates that NST and DA-TPS operate orthogonally and synergistically, effectively reconciling frame-wise accuracy with inter-frame stability.

\vspace{-0.1cm}
\section{Conclusion}

In this paper, we proposed FastVC for causal and real-time wide-angle video portrait rectification. Specifically, FastVC introduces a distortion-aware TPS (DA-TPS) module for efficient geometric modeling by aligning non-uniform control points with physical lens priors. Furthermore, we developed a dual-level spatiotemporal learning scheme that integrates neighborhood spatiotemporal (NST) attention at the model level with sparse supervision at the optimization level. Together with a specialized full-body video dataset, these synergistic components empower FastVC to effectively resolve complex lens distortions while maintaining high temporal stability. Extensive experiments demonstrate that FastVC achieves state-of-the-art performance while maintaining high computational efficiency for mobile deployment.



\nocite{ma2019effective, fan2022wide, zhang2023wide, zhang2025madcow, luan2025lifting, nie2022deep, zhao2020pwstablenet, choi2020deep, xu2024beyond, nie2023deep, guo2023content, shi2022deep, wang2023neural, zhou2024recdiffusion, wang2022point, liu2022content, peng20243d, darmon2022improving, hassani2023neighborhood, arnab2021vivit, r27}
\bibliography{fastvc}


\end{document}